\documentclass[runningheads]{llncs}
\usepackage[T1]{fontenc}
\usepackage{cite}
\usepackage{graphicx}
\usepackage{amsmath,amssymb,amsfonts}
\usepackage{algorithmic}
\usepackage{graphicx}
\usepackage{textcomp}
\usepackage{xcolor}
\usepackage{url}
\usepackage{orcidlink}
\newsavebox\myrhs
\newlength\myrhswd
\newlength\myrhssepwd
\sbox\myrhs{%
  \parbox[t]{\myrhswd}{%
    \begin{flushright} 
      This preprint has not undergone peer review (when applicable) or any post-submission
improvements or corrections. The Version of Record of this contribution is published in the Lecture Notes in Computer Science ((LNAI,volume 15561)) and is available online at https://doi.org/10.1007/978-981-96-3522-1\_26 %
    \end{flushright}%
  }%
}
\AddToHookNext{shipout/background}{
  \put(\dimexpr\paperwidth-\myrhswd-\myrhssepwd\relax,-\baselineskip){%
    \usebox\myrhs
  }%
}

\begin{document}
\title{Perception of Emotions in Human and Robot Faces: Is the Eye Region Enough?}
%
%
\author{Chinmaya Mishra\inst{1}\orcidlink{0000-0002-9223-1230} \and
Gabriel Skantze\inst{2, 3}\orcidlink{0000-0002-8579-1790} \and
Peter Hagoort\inst{1, 4}\orcidlink{0000-0001-7280-7549} \and
Rinus Verdonschot \inst{1, 5} \orcidlink{0000-0002-7124-4091}}
\authorrunning{C. Mishra et al.}
%
\institute{Max Planck Institute for Psycholinguistics, Nijmegen, The Netherlands\\
\email{chinmaya.mishra@mpi.nl}\\
\and
KTH Royal Institute of Technology, Stockholm, Sweden\\
 \and
Furhat Robotics AB, Stockholm, Sweden\\
 \and
Donders Institute for Brain, Cognition and Behaviour,
Radboud University, Nijmegen, The Netherlands
 \and
Maastricht University, Maastricht,The Netherlands\\}
\maketitle              
\begin{abstract}
The increased interest in developing next-gen social robots has raised questions about the factors affecting the perception of robot emotions. This study investigates the impact of robot appearances (human-like, mechanical) and face regions (full-face, eye-region) on human perception of robot emotions. A between-subjects user study (N = 305) was conducted where participants were asked to identify the emotions being displayed in videos of robot faces, as well as a human baseline. 
Our findings reveal three important insights for effective social robot face design in Human-Robot Interaction (HRI): Firstly, robots equipped with a back-projected, fully animated face -- regardless of whether they are more human-like or more mechanical-looking -- demonstrate a capacity for emotional expression comparable to that of humans. Secondly, the recognition accuracy of emotional expressions in both humans and robots declines when only the eye region is visible. Lastly, within the constraint of only the eye region being visible, robots with more human-like features significantly enhance emotion recognition.

\end{abstract}

\section{Introduction} \label{Intro}
There has been a surge in the development of next-generation social robots. Numerous commercial entities have proposed their versions of general-purpose robots, such as Optimus\footnote{\url{https://en.wikipedia.org/wiki/Optimus_(robot)}}, 
GR-1\footnote{\url{https://robots.fourierintelligence.com/}}, and Ameca\footnote{\url{https://www.engineeredarts.co.uk/robot/ameca/}}. While many new robots maintain a humanoid body design akin to NAO\footnote{\url{https://www.aldebaran.com/en/nao}} and Pepper\footnote{\url{https://www.aldebaran.com/en/pepper}}, the robot faces exhibit significant diversity, ranging from a highly human-like face in Ameca to a blank face design in Optimus. This calls for more research investigating how the design of the face affects the perception of social robots, and consequently, the interactions humans will have with them. 

Social robots, by definition, are designed to conduct human-like interactions~\cite{hegel2009understanding}. A key component of human communication is facial expressions which are used to convey meaning~\cite{elliott2013facial}, build relationships~\cite{lazarus2006emotions}, and help in decision making~\cite{so2015psychology}. Prior studies suggest that our brains perceive robot facial expressions similarly to human expressions~\cite{chammat2010reading, craig2010assessment}. Thus, social robots must not only recognize human emotions but also be able to convey them. Additionally, a robot can use its facial expressions to signal its intentions and internal state to a human interlocutor, which would make it easier for them to understand the robot and have a more seamless interaction with it. Modeling appropriate robot emotions is an active field of research. It has been found that robots expressing emotions are perceived as more intelligent~\cite{gonsior2011improving} and trustworthy~\cite{cominelli2021promises}. However, it is equally important to investigate the factors influencing the perception of robot emotions. Identifying these factors would help design social robots that are easier to understand and interact with. 

Researchers have investigated how robot facial expressions are perceived by humans based on robot form and appearance. An early study on emotion recognition with the Feelix robot found that adults recognize emotions in still images of the robot similarly to human faces~\cite{canamero2001show}. \cite{breazeal2003emotion} obtained similar results, indicating that individuals could interpret the robot's facial expressions from both images and videos. Literature indicates two branches of research into robots' emotion recognition by individuals depending on the robot form-factor; the first involves robots with a human-like face~\cite{danev2017development, becker2011evaluating, lazzeri2015can}, and the second involves non-humanoid robot faces~\cite{beer2010recognizing, cohen2011child}. 
This leads to the first research question:

\textbf{RQ1}: \textit{Does having a human-like face improve the recognition of a robot's emotions?}\\

The answer to this question is not clear from the literature. \cite{beer2010recognizing} investigated this query in virtual agents, comparing recognition rates for human faces, synthetic human faces, and a non-human-like virtual agent (iCat). Their results indicated a higher recognition for the human face, followed by the synthetic human face, and lastly, the virtual agent. \cite{chevalier2015impact} also reported that emotions in a female humanoid virtual agent (Mary) were better recognized than in Nao and Zeno. \cite{lazzeri2015can} and \cite{becker2011evaluating} assessed humanoid android faces against human faces. \cite{lazzeri2015can} found robot facial expressions were on par with human expressions, while \cite{becker2011evaluating} noted human emotions surpassed those of the Geminoid F robot. While these trends suggest that greater human-likeness enhances emotion recognition, this remains uncertain for human-like robot faces. Moreover, studies involving robots do not compare recognition between human-like and mechanical-looking robot faces, hindering clarity on human-likeness impact. Thus, we propose our first hypothesis:

\textbf{H1}: \textit{Human-like robot faces yield better emotion recognition compared to mechanical-looking robot faces}\\

Another aspect to consider is the role of specific face regions in emotion recognition. This stems from the broad variation in robot face designs, resulting in diverse implementations of facial regions. For instance, robots like Nao and Pepper feature static faces devoid of human-like movements, while others, such as Fuahat~\cite{moubayed2013furhat} and Ameca, possess full-face designs with human-like movements across all facial regions. This leads to our second research question: 

\textbf{RQ2}: \textit{Is it necessary to model the entire robot face with intricate human-like movements, or could we focus solely on certain regions, like the eyes?}\\

This question not only sheds light on the significance of distinct facial regions in emotion recognition but also offers a chance to simplify robot emotion generation by reducing complexities. Previous studies in psychology show the significance of seeing full-face over specific facial regions in emotion recognition \cite{baron1997there, sullivan2007age}, however, they also point to the fact that information for emotion recognition is not distributed evenly across the entire face. For example, studies suggest that the eye region alone provides sufficient information for emotion recognition~\cite{baron1997there, wegrzyn2017mapping}. \cite{baron1997there} compared the emotion recognition from pictures of the eye-region, mouth region and the full face. Their results indicated that the eye-region was as informative as the full face for complex emotions. Real-world examples include animated characters like those in the movie WALL-E (e.g., WALL-E, EVE, MO), which use minimalistic eye expressions to convey emotions and meanings\footnote{\url{https://www.pixar.com/feature-films/walle}}.  

Insights from human emotion recognition studies form the basis to investigate modeling specific face regions (like eye-region) instead of the full face for social robot design. However, this aspect remains less explored in the literature, possibly due to limited platforms with capabilities for human-like facial and eye movements. For instance, social robots like Pepper and Nao feature static eyes-only designs, precluding comparisons of emotion recognition between eye-only and full-face expressions. Some studies have tried to evaluate emotion recognition from robots' eye expressions~\cite{kang2021preliminary, barrett2019virtual} and find the best ways to model them~\cite{pollmann2019s, greczek2011using, chumkamon2014robot, barrett2019virtual}. In a study~\cite{danev2017development} on ``animated faces'' for the MASHI robot, researchers compared emotion recognition rates between full-face and eye-region expressions, finding that while the full-face yielded better recognition, eye-region expressions remained acceptable. However, these studies have been limited to either virtual characters or robots with limited expressive capabilities, such as Nao or Pepper. This leads us to our second hypothesis:

\textbf{H2}: \textit{Full face expressions will lead to better emotion recognition compared to eye-region only}\\

To explore the impact of robot appearance and facial regions on emotion recognition, we conducted a between-subjects user study, comprising two online experiments. One experiment centered on full-face emotion recognition, while the other focused solely on the eye region. In both studies, participants were tasked with identifying emotions conveyed in video recordings featuring a human, a human-like robot, and a mechanical-looking robot.

\section{Materials and Methods} \label{methods}

\subsection{Robot Platform} \label{robot}
We used the Furhat robot~\cite{moubayed2013furhat} for this study, a humanoid robot head featuring a 3D animated face projected onto a translucent mask via back projection. This setup enables the robot to adopt diverse appearances, spanning from realistic human-like to mechanical characters. Furthermore, Furhat can perform nuanced facial movements, resulting in human-like expressions. Furhat comes with a pack of pre-installed characters with a diverse range of gender, age, and human-likeness. We considered three main criteria when selecting the characters on the robot for the experiment. Firstly, we wanted to choose one robot character with a human-like appearance and another with a more mechanical look.  Secondly, we wanted similar facial expressiveness in both characters to eliminate expressiveness as an influencing factor. Finally, we needed to ensure that the robot's face was adequately visible in the video recordings shown to the participants (see Section~\ref{setup}). Since the robot's face is back-projected onto a 3D mask, many factors such as the contrast and brightness of the robot's face (character), the lighting in the room, and the camera being used to record the video become important. 

\begin{figure*}[ht]%
\centering
\includegraphics[width= 0.95\textwidth]{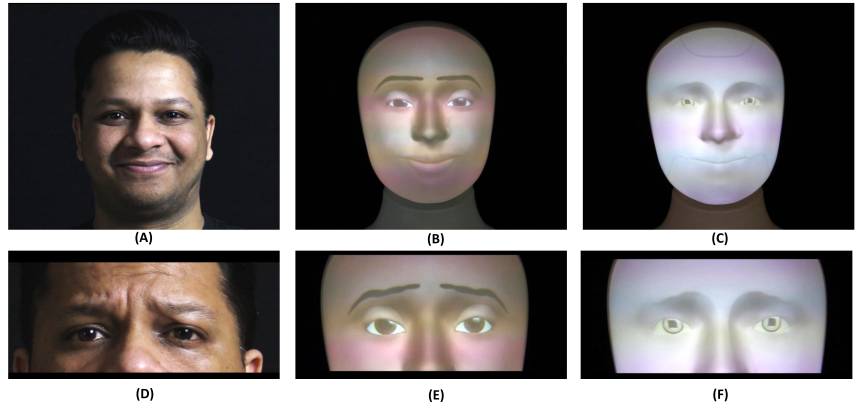}
\caption{Emotional expressions displayed by the three characters. The first row depicts the full-face expressions of \textit{Happy} by the human confederate (A), the human-like robot character Hayden (B), and the mechanical-looking character Titan (C). The second row depicts the eye-region expressions for \textit{Sad} by the human confederate (D), Hayden (E), and Titan (F)}\label{fig1}
\end{figure*}

We chose two pre-installed characters: Hayden with a realistic human-like appearance, and Titan with a mechanical look (see Fig.~\ref{fig1}). Titan gets its mechanical look from the square pupils, lack of eyebrows, white face color, and lines on the face that give the impression of its face comprising of different modular parts. The choice of mechanical look was partly due to the robotic character that was available on Furhat. Another aspect was the lack of eyebrows which is quite common in the robotic platforms being widely used in research so far, such as NAO, Pepper, Cozmo. A recent example of a highly expressive and human-like robot without any clear eyebrows is Ameca. This made the choice relevant to currently used robotic systems. Apart from these differences, Titan is able to express emotions similarly to the human-like face Hayden, as both of them share the same face model. This is in contrast to the mechanical faces that have been used in prior studies which had static eyes and mouths like Nao and Pepper. Thus, it is possible to directly compare recognition of the emotions expressed by the human-like and mechanical-looking face using the Furhat robot.

\subsection{Robot Emotions} \label{roboemotions}
Facial Action Coding System (FACS) is a comprehensive and widely used system for describing and categorizing facial expressions based on the movement of individual facial muscles~\cite{ekman1978facial}. The muscle movements, called Action Units (AUs), are assigned numerical codes to represent different facial expressions and emotions. FACS has been widely used in modelling robots' expressions in HRI~\cite{stock2022survey, beer2010recognizing, lazzeri2015can, so2015psychology, barrett2019virtual}. 

For this study, we modeled the six basic emotions~\cite{ekman1969pan} on Furhat. Since Furhat employs Apple's ARKit parameters for its face model, we mapped the FACS AUs to their corresponding ARKit parameters to generate facial expressions. Each ARKit parameter can be assigned a value between 0 (no movement of the AU) and 1 (maximum movement of the AU). For this experiment, we assigned the maximum value (i.e., 1) to the ARKit parameters to exaggerate the robot's facial expressions to make them easily recognizable in the videos. This was in line with the findings from \cite{makarainen2014exaggerating}, where the authors concluded that a robot's facial expressions should be exaggerated to communicate emotional content instead of just mimicking human facial expressions. Table~\ref{tab:AU} shows the AUs used to generate the basic emotions (adopted from \cite{clark2020facial}).

\begin{table}[ht]
\caption{Mapping of FACS AUs to emotion categories used in the study}
\label{tab:AU}
\begin{center}
    \begin{tabular}{c|c} 
    \hline
    Emotion & Action Units (AU) \\
    \hline
    Amusement/ Happy & 6 + 12\\
    Sadness & 1 + 4 + 15\\
    Anger & 4 + 5 + 7 + 24\\
    Awe/ Surprise & 1 + 2 + 5 + 26\\
    Fear & 1 + 2 + 4 + 5\\
    \hline
\end{tabular}
\end{center}
\end{table}

\subsection{Experimental Setup} \label{setup}
To evaluate our hypotheses (refer to Section~\ref{Intro}), we conducted a between-subjects user study with 3 variables (2 face regions × 3 appearances × 7 emotions). The two face regions studied were full-face expressions and eye-region-only expressions. Each group was shown either of the face regions: the full-face or eye-region-only expressions. Three appearance conditions were defined: one with expressions by a human confederate (referred to as H) serving as the control, and the other two featuring emotions expressed by the robot characters Hayden (Ha) and Titan (Ti). These robot characters represented varying degrees of human likeness, allowing us to examine their impact on emotion recognition. Instead of images, short videos were used as stimuli for the user study. This was because still images capture only a snapshot of the emotion being expressed~\cite{beer2010recognizing} and contain very little information about expressive posturing~\cite{breazeal2003emotion}. We recorded videos of 6 emotions, \textit{Happy, Sad, Anger, Surprise, Disgust,} and \textit{Fear}, for each of the face types (H, Ha, and Ti), with a baseline \textit{Neutral} expression (42 videos). 

For \textbf{H1}, which pertained to the influence of the human-likeness of the robot's face on emotion recognition, we compared the recognition of emotions expressed by a human confederate, the human-looking robot character Hayden, and the mechanical-looking robot character Titan. The expressions of the confederate and robots were recorded in high resolution using the Canon HF-G30 video camera at the university lab.
The confederate was shown examples of images and videos of facial expressions using FACS before the recording. A total of 21 videos were recorded for all the emotions and appearance conditions (7 emotions × 3 appearance types).  

\textbf{H2} aimed to compare the recognition rates between full-face and eye region-only expressions. The video recordings of the full-face expressions for all three appearance types; human, Hayden, and Titan were cropped to the eye region only. The cropped region and the proportion of the visible eye-region were kept consistent for all the videos. A total of 21 eye-region videos were extracted from the original full-face recordings.

Two online experiments were designed using the survey software Qualtrics. In the first experiment (full face condition), participants were shown short videos of the robots and the confederate on the screen and asked to select the matching emotion from the options provided on the screen. The robots and the confederate did not speak (no lip sync) and the videos had no audio. The videos first showed the robots or the confederate showing a neutral facial expression, followed by expressing an emotion and then returning to a neutral facial expression. The experiment adopted a forced-choice paradigm, requiring participants to choose one of the 7 emotions displayed as radio buttons below the video. The exact question asked was: \textit{``What emotion is being expressed in the video below?''}. Video presentation order was randomized, with a constraint to ensure that consecutive videos of the same appearance type did not occur more than twice in a row. The second experiment (eyes-only condition) followed a similar design but used the cropped eye-region videos as the stimuli. All the video durations were kept between 6-8 seconds for both experiments. Each of the experiments took roughly 7-8 minutes to finish.

\subsection{Participants and Procedure} \label{participants}

We recruited a total of 305 participants via the online survey platform Prolific (\url{https://www.prolific.com/}). The participants provided their informed consent before participating in the experiments. No personally identifiable data such as IP addresses or names were collected from the participants and the data was anonymized as soon as the participants submitted their responses. In both experiments, each participant was asked to watch 21 video clips and choose the emotion displayed in them. 

The first experiment involved 153 participants (77 males, 74 females, 2 non-binary, and 1 undisclosed), aged 18 to 59 ($M = 30.05, SD = \pm 8.23$) where they viewed videos showing the full face of the robots and the confederate. The second experiment collected data from 152 participants (76 males, 74 females, 2 non-binary) aged 19 to 54 ($M = 27.70, SD = \pm 6.66$) where they viewed videos showing only the eye region of the robots and the confederate. There was no overlap between participants in both experiments. We implemented two manipulation check questions; failing either resulted in automatic discarding of the survey response. Participants received 1 GBP upon successful experiment completion. The study has received the approval by the ethics committee of the university.

\section{Results} \label{results}
A response was counted as correct if it matched the intended emotion expressed in the video. 
RStudio (version 2024.04.2) and R software version 4.4.1 was used for the statistical analysis. 
A GLMM (Generalized Linear Mixed Model) with Gaussian distribution (lme4 package) was fit with the responses as the dependent variable. The face regions, appearances, and emotions were set as the fixed effect variables, and the participants as the random effect grouping factor. The model indicated a significant main effect of the face regions ($\chi^2$ = 69.139, \textit{df} = 1, \textit{p} $<$ 0.001), appearances ($\chi^2$ = 55.931, \textit{df} = 2, \textit{p} $<$ 0.001), and emotions ($\chi^2$ = 1953.848, \textit{df} = 6, \textit{p} $<$ 0.001) on the responses. We also observed significant interaction effects between face regions and appearance ($\chi^2$ = 20.978, \textit{df} = 2, \textit{p} $<$ 0.001), face regions and emotions ($\chi^2$ = 300.544, \textit{df} = 6, \textit{p} $<$ 0.001), and, appearances and emotions ($\chi^2$ = 791.019, \textit{df} = 12, \textit{p} $<$ 0.001) on the responses. 

Age is known to influence the recognition of both human emotions~\cite{grainger2017age} and emotions expressed by virtual agents~\cite{pavic2021age}. We split the age of the participants into 3 groups: Young Adulthood (18 - 25 years), Adulthood (26 - 44 years) and Middle Adulthood (45-59 years). A GLMM was fit with responses as the dependent variable, age group as the fixed effect variable and participants as the random effect grouping factor. The model did not show any significant effect of age groups on the responses by the participants (i.e., the recognition rates of the emotions).

\subsection{Effect of Appearances} \label{appearance}
Estimated marginal effects were computed using ``emmeans'' with the Tukey method to test pairwise comparisons. It was found that participants recognized the emotions significantly better in the human face compared to the mechanical-looking Titan ($t = 5.750$, $SE = \pm0.012$, $p< 0.001$). Emotion recognition was significantly better in Hayden than in the Titan face ($t = 7.036$, $SE = \pm0.012$, $p< 0.001$). However, there were no significant differences between the recognition rates in the human face vs. Hayden ($t = 1.286$, $SE = \pm0.012$, $p= 0.4030$).

\begin{figure}[ht]%
\centering
\includegraphics[width=\textwidth]{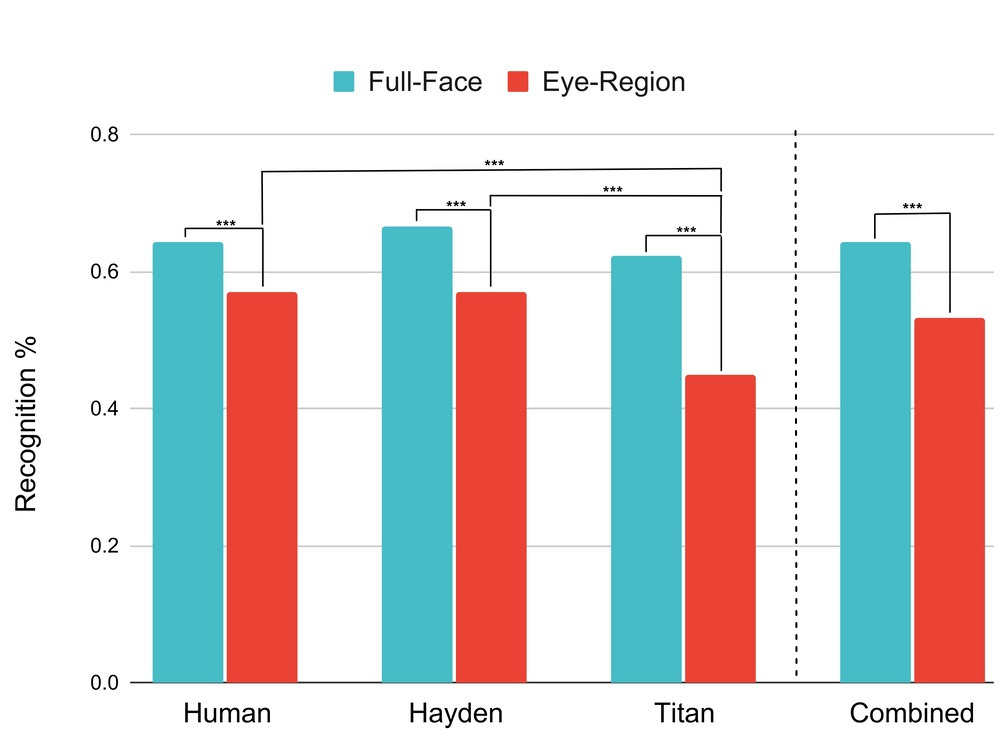}
\caption{Emotion recognition score for each of the three appearances and the combined results for both the face region conditions: full-face and eye-region. *** indicates a significant difference with $p < 0.001$}\label{fig2}
\end{figure}

However, on further analyzing the results for the interaction between appearances and face regions, it was found that these differences only held for the eye-region conditions. When looking specifically at this condition, there was a significant decrease in the recognition rate from human face to Titan ($t = 7.020$, $SE = \pm0.017$, $p< 0.001$) and Hayden to Titan ($t = 7.573$, $SE = \pm0.017$, $p< 0.001$), again in line with \textbf{H1}. However, we did not find any significant differences between the recognition rates for the full-face data based on the appearances (see Fig.~\ref{fig2}). This indicates that the significant main effect of appearance is driven by the eye-region condition.  

\subsection{Effect of Facial Regions} \label{faceEffect}
Marginal estimates with ``emmeans'' revealed that participants recognized the emotions significantly better when they were shown the full-face videos as compared to the eye-region videos ($t = 8.315$, $SE = \pm0.013$, $p< 0.001$). Overall, participants were able to recognize 64.4\% of the emotions correctly when shown the full face of the robots. Recognition was 51.3\% when only the eye-region videos of the robot were shown. Additionally, the eye-region recognition was higher than in a previous study with similar stimuli (49.1\% in \cite{barrett2019virtual}). This supports \textbf{H2}, which predicted that emotion recognition from a full face should be better than just the eye region.

\begin{figure*}[ht]%
\centering
\includegraphics[width=\textwidth]{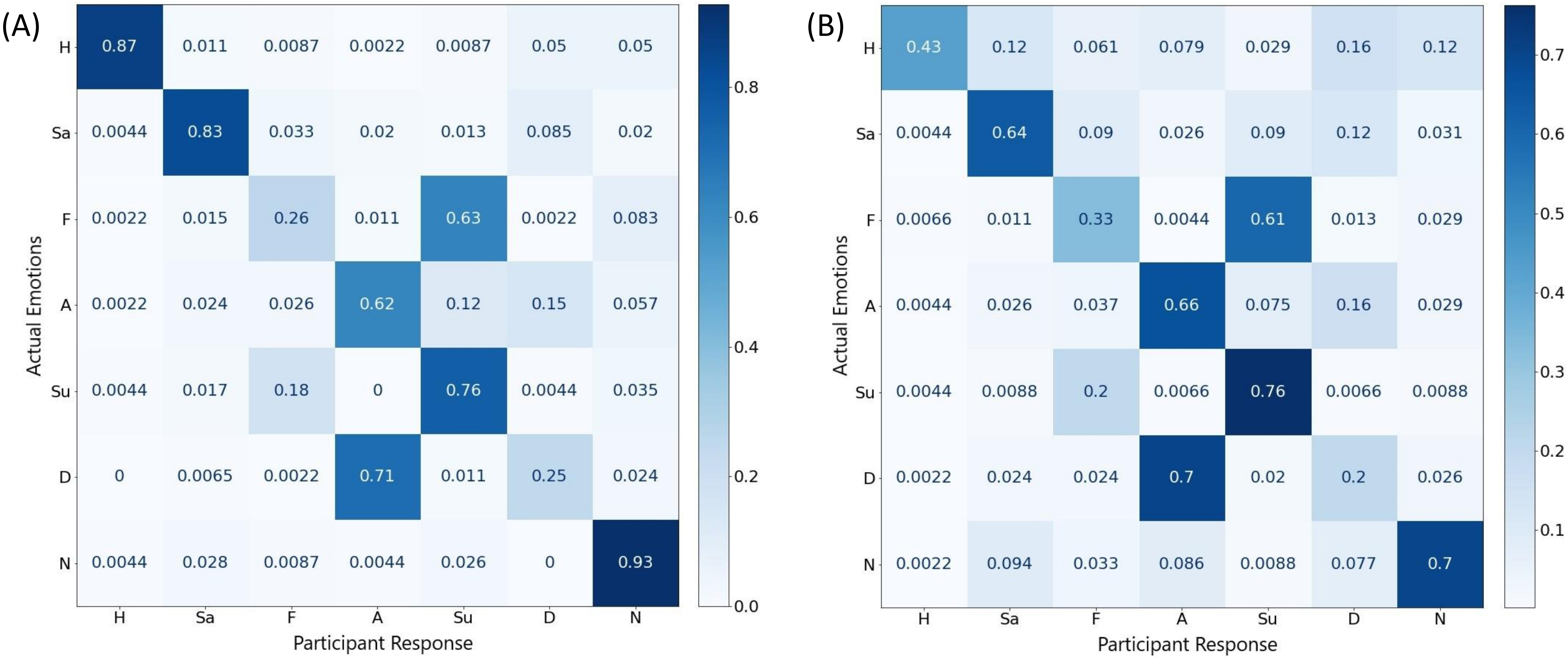}
\caption{Normalized confusion matrix between actual and selected emotions by the participants under both the face region conditions. Sub-figure (A) depicts the confusion matrix for the full-face condition and (B) depicts the confusion matrix for the eye-region condition. Emotion abbreviation in the figure: H - \textit{Happy}, Sa - \textit{Sad}, F - \textit{Fear}, A - \textit{Anger}, Su - \textit{Surprise}, D - \textit{Disgust}, N - \textit{Neutral}}\label{fig3}
\end{figure*}

Further pair-wise comparisons between face regions and emotions were conducted. Figure~\ref{fig3} shows the confusion matrix with recognition accuracy for both face region types. It was observed that the recognition rates for \textit{Fear, Anger, Surprise,} and \textit{Disgust} were similar for both full-face and eye-region videos. Significant differences were found for \textit{Happy} ($t = 15.986$, $SE = \pm0.027$, $p< 0.001$), \textit{Sad} ($t = 6.743$, $SE = \pm0.027$, $p< 0.001$), and \textit{Neutral} ($t = 8.314$, $SE = \pm0.027$, $p< 0.001$), with higher recognition rates in full-face videos. 

\section{Discussion} \label{discuss}
This study aimed to explore how robot appearances and facial regions affect robot emotion recognition. GLMM results highlighted the significant effects of appearances and facial regions on emotion recognition. Analysis of estimated margins 
indicated that greater human-likeness correlated with improved emotion recognition. Notably, emotion recognition rates showed no significant differences between the human and human-like robot faces (Hayden) in full-face videos, which further strengthens the advantage of a human-like face design. However, this difference only holds when the perception is limited to the eye region condition which means that \textbf{H1} is only supported when the full face is not visible. This suggests that emotions expressed by a robot having a full-face (animated) design, irrespective of its appearance (human-like or mechanical), are recognized similarly to those expressed by a human signifying the advantage of having a fully-animated robot face. One reason for the lack of significant difference between recognition rates in the full-face condition could be because, even though Titan is mechanical-looking, it still has a very expressive mouth region. In comparison with mechanical-looking robots like iCub, Furhat's Titan character appears more human-like when the full face is viewed. As mentioned in Section~\ref{robot}, we acknowledge that the specific design of the mechanical-looking robot used here is quite arbitrary (in our case, having an expressive mouth but no eyebrows), so the results might not generalize to all mechanical-looking robots however it aligns with the robotic platforms being widely used in research so far, such as NAO, Pepper, Cozmo.

On the other hand, a significant difference was found in emotion recognition favoring human-likeness for eye-region videos. This can be attributed to the video cropping, which obscured human-like features and emphasized mechanical aspects (see Section~\ref{robot}). The significant decrease in emotion recognition between full-face and eye-region for Titan's emotions is indicative of the same. A comparison of Hayden and Titan's eye-region videos sheds light on how mechanical appearance impacts emotion recognition. This raises questions about the role of eyebrows or pupil shape in the difficulty in recognizing Titan's emotions and whether having a mouth mitigates these effects, as full-face emotion recognition did not differ significantly. 
These questions warrant further exploration in broader studies. Our findings underscore the potential of human-like robot appearance for enhancing emotion recognition. Future research should explore diverse robot embodiments to address appearance variability among robot designs. 
In line with \textbf{H2}, the recognition rate for full-face was significantly higher than for the eye-region videos.  Pair-wise analysis of estimated margins also supported this hypothesis, with significantly higher recognition for 3 emotions (\textit{Happy, Sad, Neutral}) in the full-face videos. This is in line with previous findings which reported better recognition with full-face stimuli~\cite{danev2017development}. However, it is worth noting that the difference in recognition for four emotions between eye-region and full-face responses was not statistically significant. This could point to the capability of the eye-region to express emotions sufficiently depending on the emotions that need to be expressed. Nonetheless, omitting a full face may result in the loss of valuable additional cues that could greatly help in emotion recognition. For example, we observe a significant decrease in the recognition of \textit{Happy} when moving from full-face to eye-region. This could be attributed to the fact that the major cues for happiness lie in the mouth region~\cite{wegrzyn2017mapping}. This needs to be kept in mind when deciding whether or not to model the full face when designing a social robot's face.

It was found that participants struggled to recognize \textit{Fear} and \textit{Disgust} for both facial region types (see Fig.~\ref{fig2}), consistent with findings from a study using a virtual eye region model~\cite{barrett2019virtual}. Additionally, participants often confused \textit{Fear} with \textit{Surprise} and \textit{Disgust} with \textit{Anger}. This could be explained by the Perceptual-Attentional Limitation Hypothesis which posits that the confusion between these emotions arises due to their shared muscle movements and visual similarities~\cite{roy2015confusion, hendel2023exploration}. 

This study is aimed as a first step at investigating the overall influence of appearance and facial regions (if any) on emotion recognition of robot expressions. Focusing on the specifics such as the influence of the features such as eyebrows or pupil size will be looked into in future studies and was beyond the scope of the present paper.

\section{Conclusion} \label{conclusion}
In this study, we investigate the influence of appearance and facial region on robot emotion recognition, with specific attention to the human-likeness of their appearance and the role of the eye-region. A comprehensive between-subjects user study was conducted with 305 participants. Results indicated that human-likeness improved participants' ability to recognize emotions in robots. Additionally, recognition rates from the eye-region, while not as effective as full-face, were found to be within a comparable range. However, it is essential to acknowledge that foregoing the modeling of the full face may result in the loss of crucial cues for certain emotions, as exemplified by the significance of mouth cues in recognizing happiness. Based on the findings of our study we propose the following overall recommendations when designing robot faces and modelling emotions on them:

\begin{itemize}
    \item A back-projected fully animated face (regardless of whether it is more human-like or more mechanical-looking) has similar capabilities to express emotions as a human face.
    \item If possible, the whole face should be modeled instead of just the eye region to avoid losing emotional expressiveness.
    \item If a robot face is designed with only the eye region, having more human-like features is important for better emotion recognition.
\end{itemize}


\section{Ethical Impact Statement}
This study involved measuring the perception of the facial expressions exhibited by robots by the participants and trying to see if the human-likeness and facial regions of a robot's face had any influence. Two key considerations need to be taken while generalizing results from this work. Firstly, the work focuses on the recognition rates of only the 6 basic emotions whereas humans exhibit far more than just these emotions. Moreover, human emotion is multi-modal and context-dependent. This study only looked at the recognition rates of emotions when expressed on the face without any context or associated cues from other modalities such as speech. This restricts the interpretation of the results to only one modality -- visual and also just the 6 basic emotions. Secondly, the study only used a Furhat robot which ties the results to the capabilities and implementation of expressiveness on a single robot platform. To generalize the findings, further studies with a wider selection of robots and appearances need to be carried out. Nonetheless, this study provides a first step toward assessing the influence of the appearance of a robot on emotion recognition by humans. 

\begin{credits}
\subsubsection{\ackname} This project has received funding from the European Union's Framework Programme for Research and Innovation Horizon 2020 (2014-2020) under the Marie Skłodowska-Curie Grant Agreement No. 859588.

\subsubsection{\discintname}
Authors CM and GS were employed by Furhat Robotics AB. The authors have no competing interests to declare
that are relevant to the content of this article.
\end{credits}
%
%
%
\bibliographystyle{splncs04}
\bibliography{paper}
%




\end{document}